\documentclass[runningheads]{llncs}
\usepackage[T1]{fontenc}
\usepackage{cite}
\usepackage{graphicx}
\usepackage{geometry}
\usepackage{hyperref}
\usepackage[dvipsnames]{xcolor}
\begin{document}
\title{Personalised 3D Human Digital Twin with Soft-Body Feet for Walking Simulation}
\titlerunning{PERSONALISED 3D HUMAN DIGITAL TWIN}
%
\author{Kum Yew Loke\inst{1} \and
Sherwin Stephen Chan\inst{1,2} \and
Mingyuan Lei\inst{2}\and
Henry Johan\inst{2}\and
Bingran Zuo\inst{2}\and
Wei Tech Ang\inst{1,2}}
\authorrunning{K. Y. Loke    et al.}
%
\institute{School of Mechanical and Aerospace Engineering, Nanyang Technological University, Singapore 
\newline
\email{lokekumyew@gmail.com},
\email{sherwin.chan@ntu.edu.sg}
\and Rehabilitation Research Institute of Singapore (RRIS), Singapore
}
\maketitle              

\begin{abstract}
With the increasing use of assistive robots in rehabilitation and assisted mobility of human patients, there has been a need for a deeper understanding of human-robot interactions particularly through simulations, allowing an understanding of these interactions in a digital environment. There is an emphasis on accurately modelling personalised 3D human digital twins in these simulations, to glean more insights on human-robot interactions. In this paper, we propose to integrate personalised soft-body feet, generated using the motion capture data of real human subjects, into a skeletal model and train it with a walking control policy. Through evaluation using ground reaction force and joint angle results, the soft-body feet were able to generate ground reaction force results comparable to real measured data and closely follow joint angle results of the bare skeletal model and the reference motion. This presents an interesting avenue to produce a dynamically accurate human model in simulation driven by their own control policy while only seeing kinematic information during training.
\end{abstract}
\section{Introduction}
Due to the onset of an ageing population around the world, there has been a rise in the number of people with mobility impairments. As advancements in robot-assisted mobility systems have enhanced the mobility of these people~\cite{ref_1}, significant efforts have been put into developing these technologies. One of the challenges faced in these efforts is the time required for human trials, due to the need for approval from the Institutional Review Board~\cite{ref_2} and time for recruiting patients. Additionally, cost and patient availability are other concerns that come with human trials. This bottleneck in validating new developments may be helped by human-robot interaction simulations that have high-fidelity human models.

The human models for existing simulations appear to fall within two categories. The first is simulation models that focus on modelling human motion typically with a skeletal (SK) or musculoskeletal (MSK) model, for example, AnyBody~\cite{ref_4} and OpenSim~\cite{ref_5}. Despite the proven accuracy in modelling human motion, there is often a lack of high-fidelity contact modelling for these models, as seen from the implementation by Christensen \textit{et al.} ~\cite{ref_6},  with simplified contact definition. The other category focuses on modelling interaction and shapes, a key example being RCareWorld~\cite{ref_7}. Despite being the only simulator with a full-body human model with soft-body modelling, RCareWorld still appears to have limitations. Its human model is in predefined configurations (one caregiver and six care recipients). Shape and mobility parameters are fixed for each model, lacking the ability to personalise the human model to individuals. Hence, it is clear that most of these simulations lack the ability for accurate contact modelling, especially involving soft tissues. For the technologies that include soft-body simulations, their implementations lack personalised human models. This limits their use in applications involving simulations of individual subjects.

As human models for such simulations do not appear to have both personalised human shape representation and soft-body simulations, this affirms the work conducted for this paper, in generating a personalised 3D human digital twin with soft-body feet for walking simulation using motion capture data, with good ground reaction force (GRF) and joint angle profiles. To understand the best approach in generating the 3D human shape for use in this paper, reference was taken from existing research comparing commonly employed models for human shape modelling~\cite{ref_11}. From this, we concluded that SMPL is the best choice. Common soft-body simulation techniques like Extended Position Based Dynamics (XPBD)~\cite{ref_xpbd} were also evaluated against MuJoCo's approach to simulating soft-bodies with a hyperelastic model. Although approaches like XPBD show promising results, ultimately MuJoCo's approach was chosen as it was an inbuilt implementation within the MuJoCo engine. From our review, we also concluded that the approximation of soft human tissue using hyperelastic materials is viable and supported by existing research~\cite{ref_26}, proving the viability of MuJoCo's approach.

\section{Our Proposed Approach}
\begin{figure}
    \centering
    
    \includegraphics[width=0.95\linewidth]{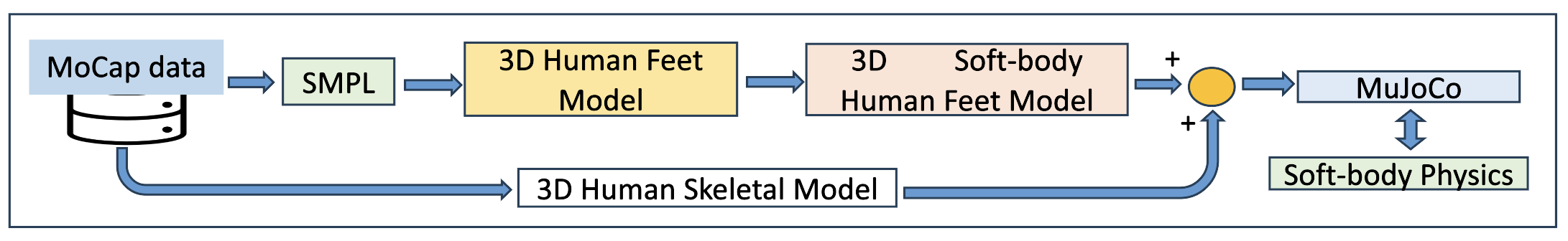}
    \caption{Simulation Pipeline}
    \label{fig:pipeline}
\end{figure}

Figure \ref{fig:pipeline} details the overall pipeline for the paper. Prior to this paper, the RRIS team implemented a pipeline where motion capture data~\cite{ref_mocap} is used to generate a personalised 3D human skeletal (SK) model~\cite{ref_beth} which is then used in simulations. With MuJoCo as the physics engine, this model is trained with a walking control policy \cite{ref_rris}. These efforts are represented in the bottom part of the figure. The top part of the figure represents the efforts in this paper. The primary aim is to use motion capture data to generate personalised 3D human foot models using the SMPL model~\cite{ref_smpl} and create soft-body objects from the 3D human foot shapes and simulate them as MuJoCo's flex objects. These flex objects will be integrated with the SK model for walking simulation.
\subsection{Generating the Feet Model}
\begin{figure}
    \centering
    \includegraphics[width=0.3\linewidth]{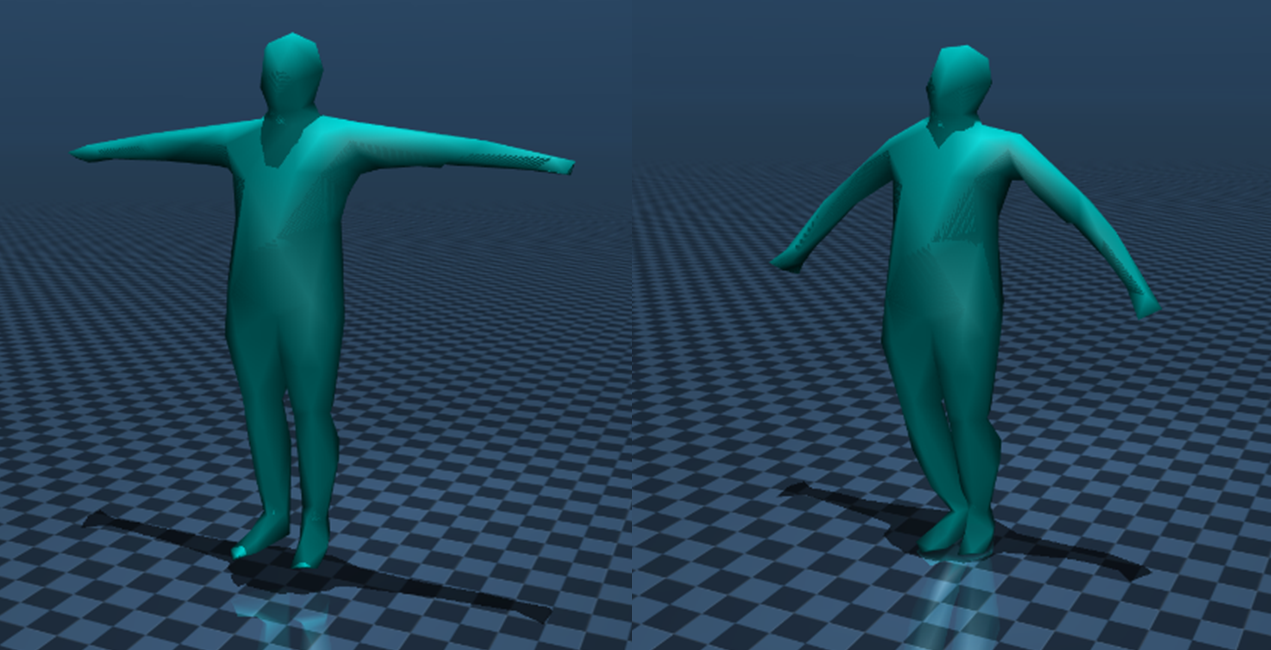}
    \includegraphics[width=0.25\linewidth]{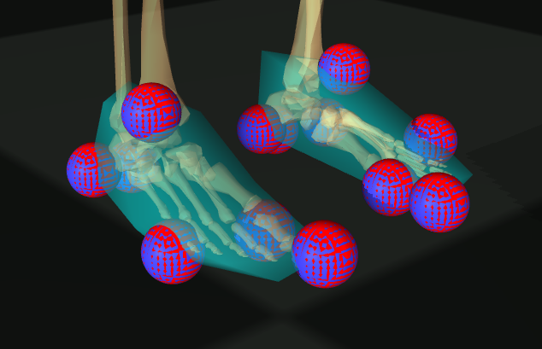}
    \includegraphics[width=0.6\linewidth]{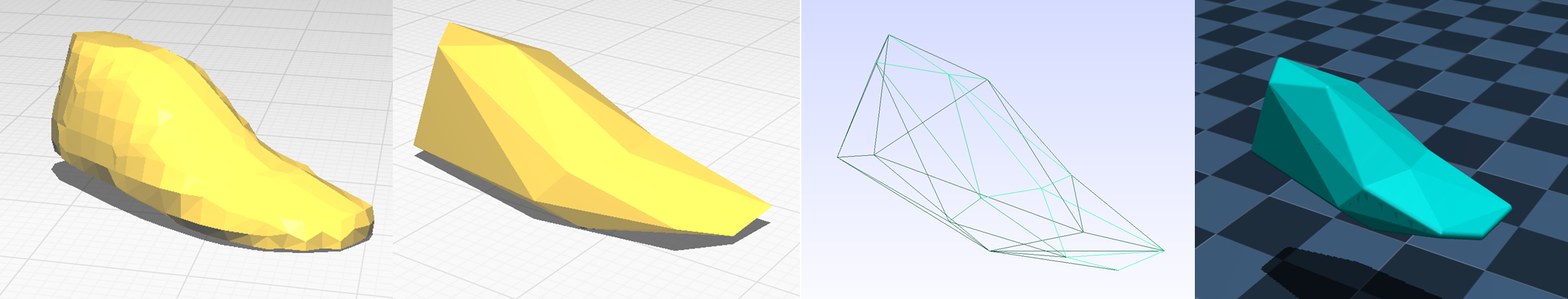}
    \caption{Top row: Full body flex model with reduced resolution (Left) and Pin connections between the flex feet and the SK model, as shown by the spheres (Right). Bottom row: From left to right, Original high-resolution foot shape, Foot shape of reduced resolution, Mesh model of the foot, flex object in MuJoCo of the Foot}
    \label{fig:fullbody-pin-foot}
\end{figure}

Originally, a full-body human model with reduced complexity, as seen in Figure \ref{fig:fullbody-pin-foot}, was generated to test the capabilities of MuJoCo, but this was too computationally intensive. Hence, as only the feet are in contact with the environment in simulations of the walking motion of a human model, we decided to focus on modelling the feet as soft-bodies. 
During testing, it was noted that there is a need to simplify the original high-resolution foot shapes to reduce computational cost. The original shapes were simplified in Blender, from around 350 vertices to 15 vertices. The Remesh Modifier (Smooth) was used to create an even surface mesh before the resolution was reduced with the Decimate Modifier. Once generated, the simplified foot shapes are converted into mesh files and loaded into MuJoCo as flex objects, as shown in Figure \ref{fig:fullbody-pin-foot}. Based on material properties of common human tissues like connective tissue, muscle, and fat~\cite{ref_22}, appropriate material properties of the flex objects are defined.
\subsection{Integrating Soft-body Feet with the Skeletal Model}
With the soft-body feet generated as flex objects, there is a need to attach the flex to the SK model. Nesting the flex objects in the existing SK model, under the skeletal part to which it is connected produces the best results, as the flex objects can follow the motion of the parent skeletal part with minimal issues.  Additionally, connections distributed through the flex performed well, ensuring that the flex translated and rotated along with the SK model. Amongst the MuJoCo connection options, Pins performed the best as welds and connects in MuJoCo are soft contacts ~\cite{ref_9}, resulting in oscillation between the flex feet and the SK model. This caused difficulty in training the walking control policy. Pins, being hard contacts, cause no oscillation of the connection and help improve the stability of the model, allowing it to be trained. The distribution of the Pin connections in the flex foot can be seen in Figure \ref{fig:fullbody-pin-foot}.
\subsection{Configuring the Material Properties of the Soft-body Feet}
\subsubsection{Material Property Values}
With the soft-body feet generated and attached properly to the SK model, we need to define appropriate material properties for the feet model to approximate human soft tissue. Literature review of human soft tissue material properties~\cite{ref_22,ref_27,ref_28,ref_29,ref_30} revealed that the material properties of human connective, muscle, and fat tissues fell within the ranges shown in Table \ref{tab_matProp}. 
\begin{table}
\centering
\caption{Range of Material Properties of Human Soft Tissues}\label{tab_matProp}
    \setlength{\tabcolsep}{6pt}
    \scriptsize
    \begin{tabular}{|l|l|l|}
    \hline
    Tissue Type&  Young’s Modulus/ kPa& Poisson Ratio
    \\ \hline 
 Connective Tissue& $1.5\times10^6$ – $2.25\times10^5$&0.3
    \\\hline
    Muscle&  164 – 39& 0.493 – 0.3
    \\ \hline 
    Fat&  18 – 24& 0.5 – 0.13
    \\\hline
    \end{tabular}
\end{table}
As the feet will be simulated as a flex object with uniform material properties, it is important to define it to represent the soft tissues in the feet as a whole. Additionally, with a relatively large range of material properties for human soft tissues, and the lack of literature specifically studying the distribution of soft tissues in the feet, it was decided that flex material properties within the range defined in Table \ref{tab_matProp} should be experimented with. The model is then personalised by fine-tuning the material property values using GRF and joint angle values to evaluate the results. As the foot has a relatively high composition of muscle and connective tissues, the flex feet were first defined with properties within the range of muscle tissue.
\subsubsection{Use of Elasticity Plugin}
In investigating approaches to inputting the material properties into the flex, two methods were utilised. The first is with the elasticity plugin, while the second is by directly defining the properties of the connections within the flex object. Experimentation with both approaches was conducted. With the elasticity plugin, the model receives inputs on the Young’s Modulus, Poisson ratio, and damping coefficient. For the damping coefficient, there is limited literature on the damping coefficient of human tissue. The value of 0.1 was chosen after experiments in simulation to determine which value reduced oscillations and generated representative behaviour.

For the method of directly modifying the flex properties, the model took two values, stiffness and damping, which defined the behaviour of the connections within the flex. Appropriate values for stiffness and damping, correlating to the actual Young’s Modulus, Poisson ratio, and damping coefficient, were determined by comparing the behaviour of flex objects defined with both approaches. Overall, the soft-body behaviours of flex defined with both methods are similar, with the elasticity plugin approach seeming slightly more realistic.
During the experimentation, it was also noted that each method had advantages and disadvantages, as detailed in Table \ref{tab_ElasProCon} below.
\begin{table}
\centering
\caption{Properties of Both Approaches in Defining Material Properties of Flex, Directly Modifying Flex Properties Versus Using Elasticity Plugin}\label{tab_ElasProCon}
\scriptsize
\setlength{\tabcolsep}{6pt}   
\begin{tabular}{|p{1.8cm}|p{4.5cm}|p{4.5cm}|}
\hline
Factors for Consideration&  Directly Modifying Flex Properties&Using Elasticity Plugin
\\ \hline 
 Properties& Define the stiffness and damping& Define the Young’s Modulus, Poison ratio and damping coefficient\\ \hline 
 Computation& Simpler computation, Able to maintain 0.001s timestep& More computation resources required; With higher stiffness, smaller timestep needed for model to be stable, usually 0.0005s or smaller\\ \hline 
 Results& Relatively accurate at approximating material properties, especially at small and large deformation& Overall, a more accurate approximation of soft properties\\\hline
\end{tabular}
\end{table}
Upon further testing and training of the control policy, we also noted that models using the elasticity plugin approach were increasingly unstable when flex objects of higher stiffness were defined, requiring much smaller timesteps to maintain stability in training and reducing its viability. Hence, in the next stages of the paper, the approach without the elasticity plugin was utilised. 

For all the trainings, 16 CPU cores were used on an AMD threadripper. With the current approach of directly defining the material properties of the flex, the model took on average 11.7 seconds to train and around 4200 iterations to converge. In comparison, the base skeletal model required 5.3 seconds per iteration and converged after 4200 iterations. As for the elasticity plugin approach, it took around 4400 iterations to converge and around 17.5 seconds per iteration.

\section{Results and Analysis}
\subsection{Walking Control Policy Training for the Skeletal Model with Soft-body Feet}
With the new models with flex feet generated, they were then trained with their walking control policies. To benchmark the performance of the control policy, results from these new models were compared with the original data captured during the motion capture session and the results from the original SK model. GRF and the joint angle values were used as metrics for dynamic and kinematic analysis respectively.

From the initial results of the training, the material properties of the flex and the position of the flex foot were fine-tuned to improve the GRF and joint angle results. This was done by progressively increasing the stiffness of the flex material and evaluating the GRF results based on the approaches detailed below. Material properties that produced better GRF results were deemed to be more representative properties. The final version of the model, model E, has flex feet of increased stiffness and position adjusted, with more distance between the bottom of the left flex foot and the SK foot. Table \ref{tab_simMatProp2} summarises the material properties of all the models.
\begin{table}
\centering
\caption{Material Properties of Models A, B, and E, with Values of Parameters Used to Define Equivalent Properties When Directly Modifying the Flex Properties}\label{tab_simMatProp2}
\scriptsize
\begin{tabular}{|p{1cm}|p{3cm}|p{3cm}|p{2cm}|p{2cm}|}
\hline
 & \multicolumn{2}{|c|}{Equivalent Human Soft Tissue Properties}&\multicolumn{2}{|c|}{Directly Modifying the Flex Properties}
 \\\hline
Model&  Young’s Modulus/kPa&Poisson Ratio& Stiffness& Damping\\ \hline 
 A& 57& 0.3& 2000&100\\\hline
 B& 110& 0.3& 5000& 100
\\ \hline 
 E& 500& 0.3& 12000& 100
\\\hline
\end{tabular}
\end{table}
\subsection{Ground Reaction Force (GRF)}
GRF serves as an important metric for dynamic analysis of the simulation results. The GRF readings collected by the force plates during the motion capture recording sessions provide a benchmark for the shape and force magnitudes of the ideal GRF to compare the GRF outputs from the simulation. 

Plots of the vertical GRF against the gait percentage are generated for four gait cycles, and these results are overlaid against the actual GRF data, with the mean and standard deviation of the actual data represented as a blue line and shaded blue area respectively. 

Additionally, the percentage of the gait cycle in which the simulation GRF lies within the standard deviation of the recorded data is calculated and displayed as the experimental match (EM) metric. This EM metric quantitatively measures how closely the simulation GRF outputs match the recorded GRF data. This method of data visualisation and comparison was inspired by a paper on training human MSK models with reinforcement learning~\cite{ref_34}. 

GRF data analysis was conducted for the left foot only due to the distribution of the GRF data collected during the motion capture sessions. Due to the placement of the force plates, there were more complete GRF recordings of the left foot compared to the right foot. With six complete left foot recordings, it was sufficient to determine a mean and standard deviation. The right foot only had three complete recordings which was insufficient to determine an accurate standard deviation. 
\begin{figure}
\centering
    \includegraphics[width=0.28\linewidth]{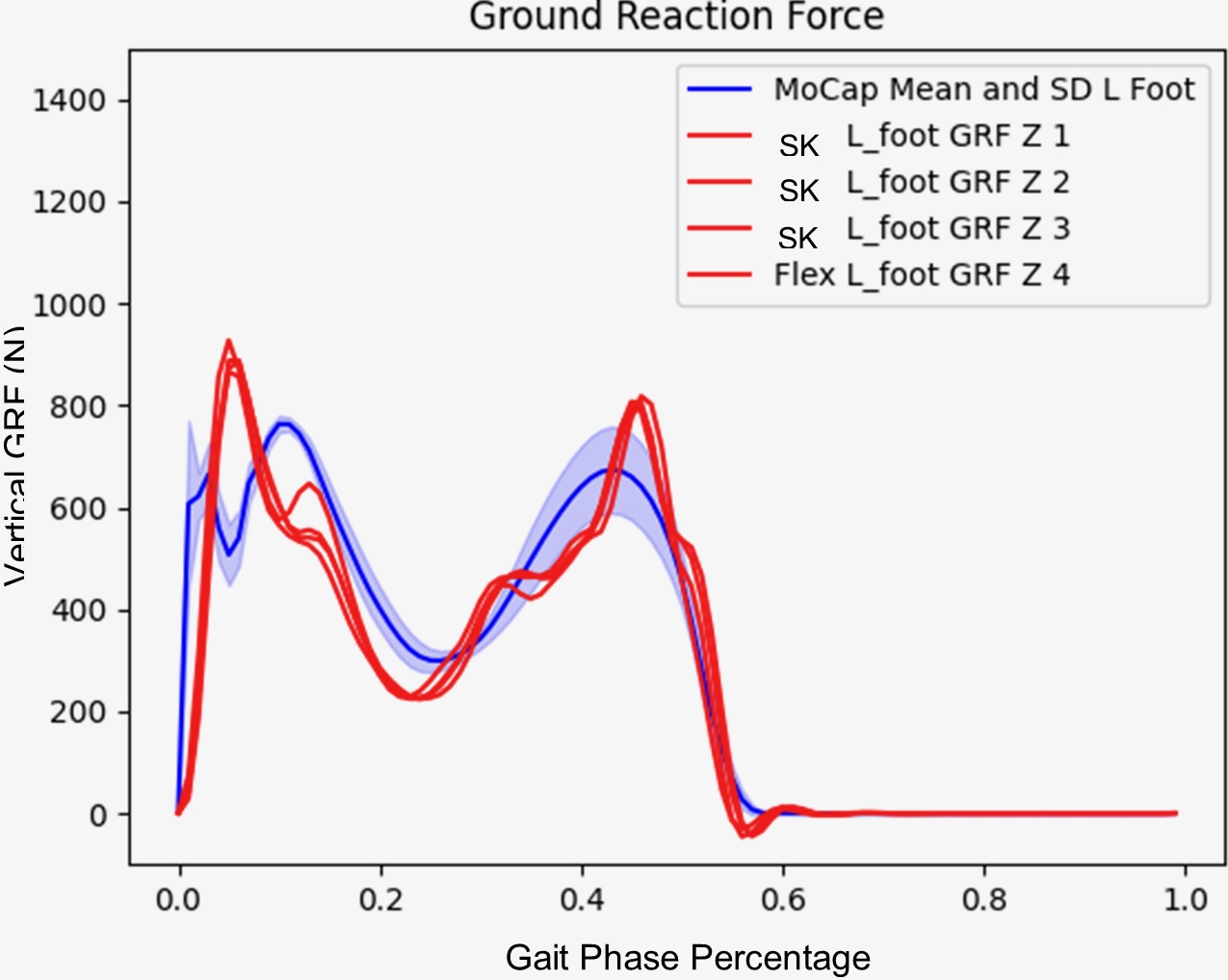}
    \space\space\space\space\space\space\space\space\space\space\space\space
    \includegraphics[width=0.28\linewidth]{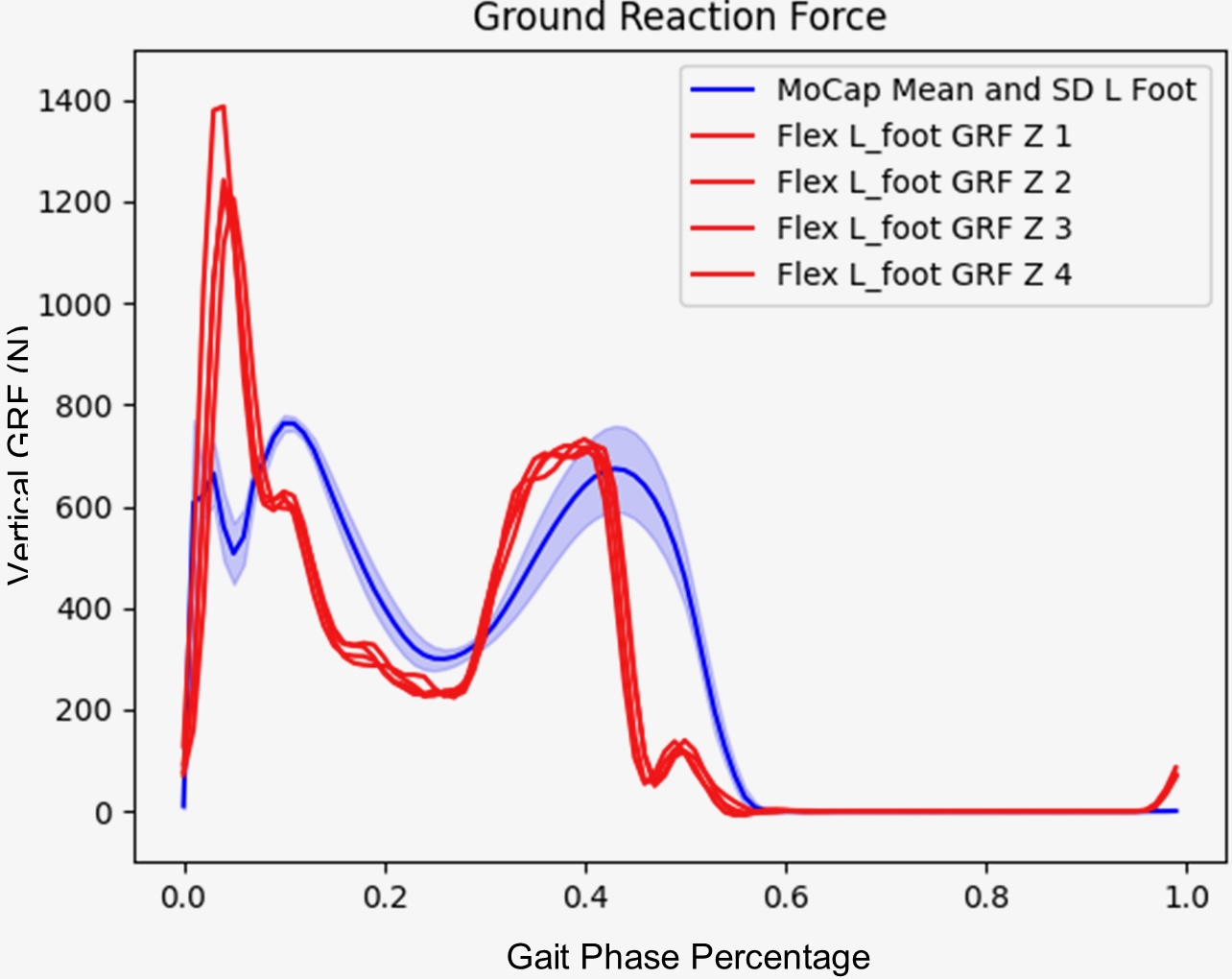}
    
    \includegraphics[width=0.28\linewidth]{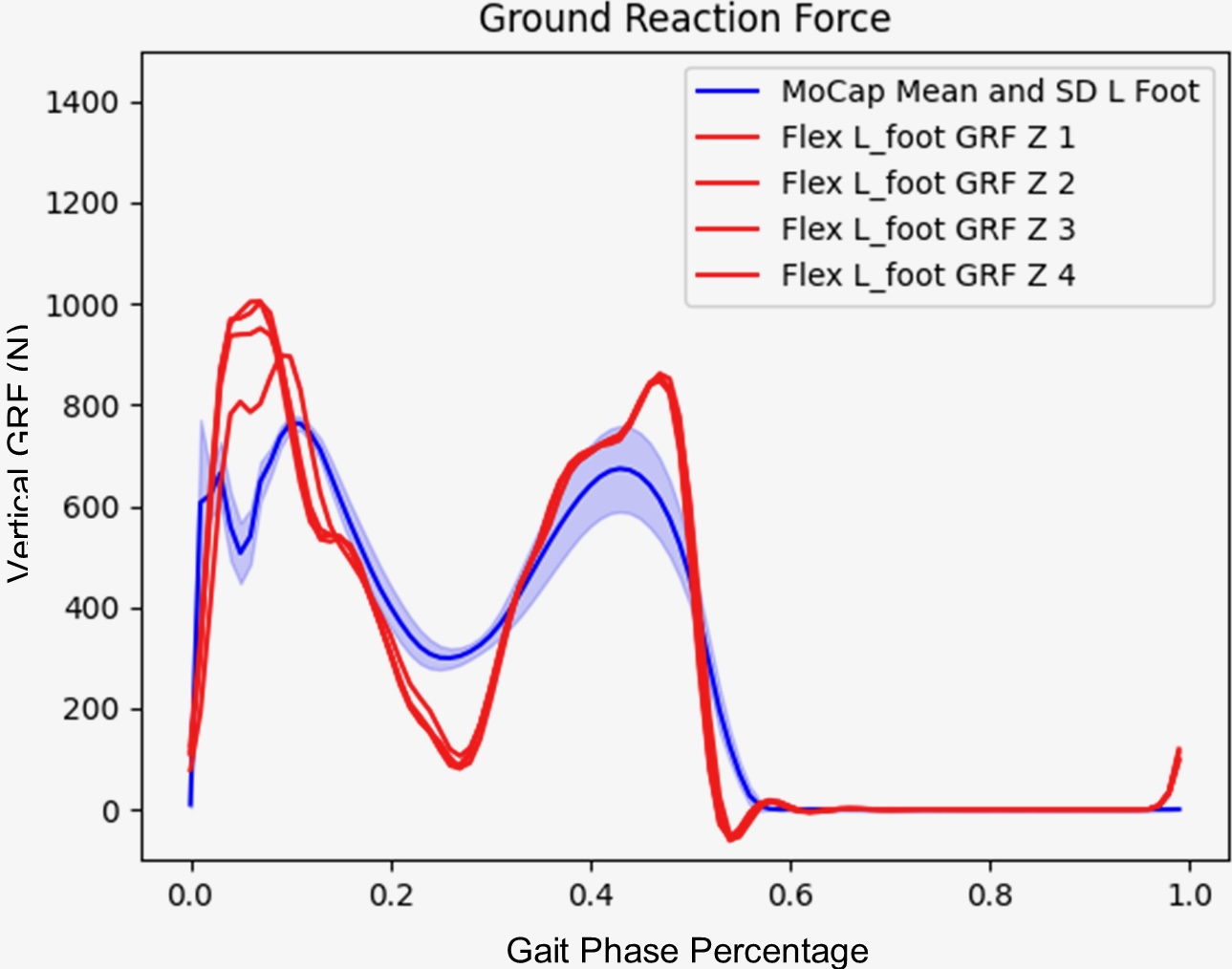}
    \space\space\space\space\space\space\space\space\space\space\space\space
    \includegraphics[width=0.28\linewidth]{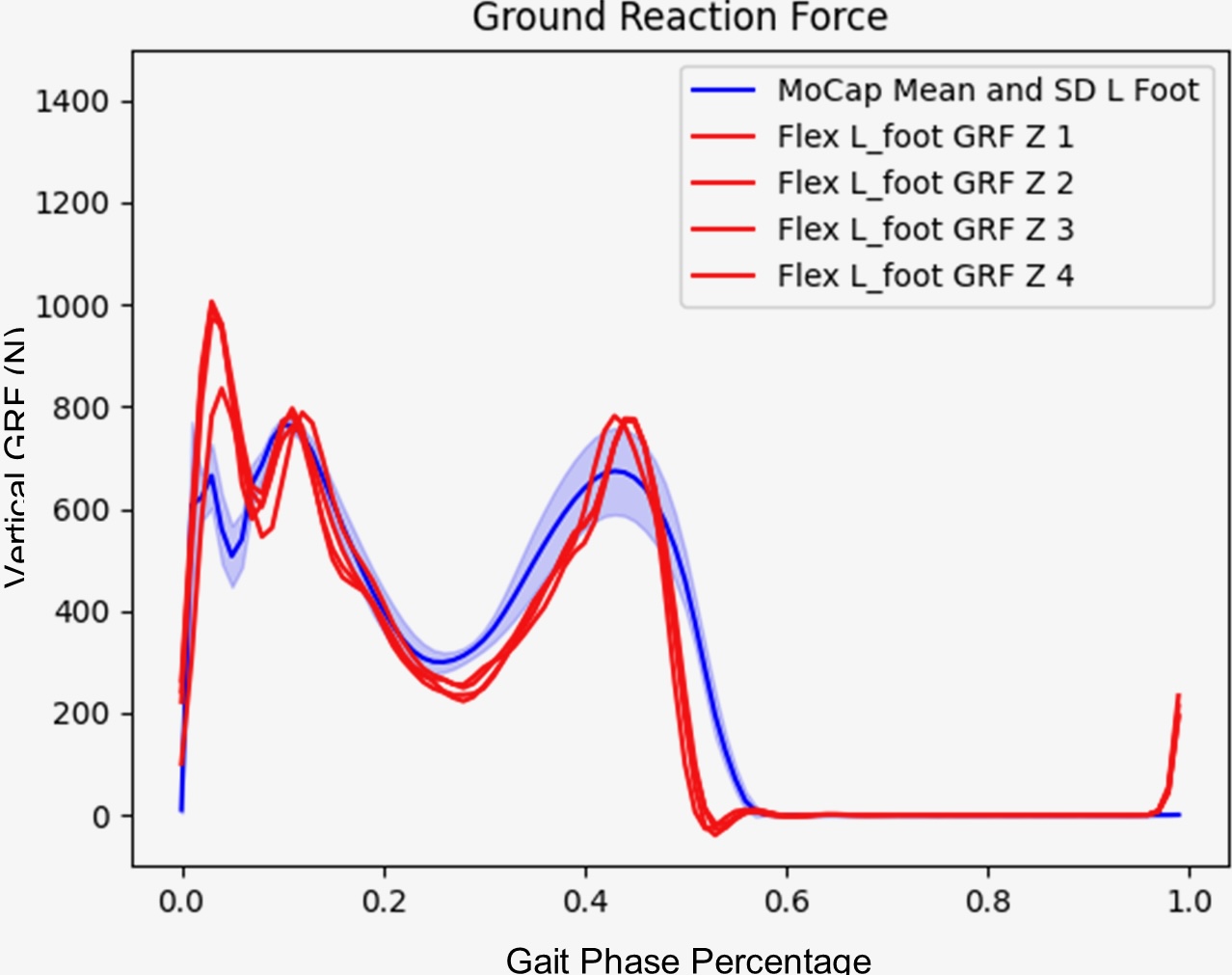}
    \caption{Plot of vertical GRF against gait percentage for models SK, A, B, and E (Top row left to right, bottom row left to right). Note that for models A, B, and E, offsets were applied to better match the GRF profiles.}
    \label{fig:1}
\end{figure}
The models used were the base SK model (model SK), the simplified flex foot with a material property at around the mid-range of muscle material properties (model A), a model with flex feet of material property at the higher-range of muscle stiffness values (model B), and the final model with flex feet of material properties in the range of muscle-connective tissues (model E). The vertical GRF plots can be seen in Figure \ref{fig:1} , as well as the EM values for all the models in Table \ref{tab_EM_SKABE}. For the GRF plots of models with flex feet, a slight offset was applied, as there was a phase difference from the results of these models, which may be explained by slight differences in the shape of the flex feet as compared to the actual soft tissue of the subject’s feet, likely due to the simplification of the flex shape and approach in attaching the flex feet to the SK model. This offset might have caused the initial gradual increase in the GRF recorded instead of the rapid increase from the start as seen in the actual GRF data.

This issue with the phase difference is also noted in the paper where the EM method of analysis was referenced from~\cite{ref_34}, where it was mentioned that even plots with profiles that match the actual natural profile could return a low EM value if there was a phase difference or offset. The GRF plot of the original model SK did not have offset applied, as the initial rapid increase in vertical GRF values is already relatively aligned between simulation and recorded data, further justifying the hypothesis that the offsets seen are a result of adding the soft-body flex feet.
\begin{table}
\centering
\caption{Experimental Match (EM) Values of Models SK, A, B, and E}\label{tab_EM_SKABE}
\scriptsize
\setlength{\tabcolsep}{2pt}
\begin{tabular}{|l|l|l|l|l|l|}\hline
 & \multicolumn{5}{|c|}{EM}\\\hline
Model Type&  Gait Cycle 1&Gait Cycle 2& Gait Cycle 3& Gait Cycle 4& Average\\ \hline 
 SK& 0.58& 0.56& 0.54& 0.55& 0.558±0.022\\\hline 
 A& 0.45& 0.46& 0.47& 0.47& 0.463±0.013\\\hline
 B& 0.55& 0.54& 0.56& 0.58& 0.558±0.022\\\hline
 E& 0.67& 0.63& 0.61& 0.60& 0.628±0.042\\\hline
\end{tabular}
\end{table}
The GRF plot of models A and B are more similar to the actual GRF plot compared to model SK, specifically in following the “M”-shaped profile. The higher EM values of model A over model SK confirm this. However, there are sharp spikes in the GRF profile of models A and B, at around 0.1 and 0.45 of the gait cycle.  These spikes in the GRF plots can be attributed to the skeletal model contacting the floor at that instance, resulting in a spike in the GRF recorded. The GRF profile for models A and B are also not smooth with secondary peaks. This is likely due to intermittent contact of the SK model with the floor, resulting in spikes in the GRF recorded and deviations from the smooth profile. These issues of contact between the skeleton and the floor needed to be limited to ensure a smoother GRF plot without spikes. 

Further increasing the stiffness of the flex model E resolved this, reducing deformation of the flex, allowing the flex to account for most of the contact with the floor and reducing the instances of the skeleton contacting the floor. The GRF profile for model E matches the actual GRF plot more closely than previous models, with a clear increase in the EM value. Hence, increasing the flex material properties' stiffness generated more accurate and representative GRF results. The video of simulation animations for the four models is \href{https://youtu.be/_HkKbqYMRdo}{linked here}.



\subsection{Joint Angles}
\begin{figure}
    \centering
    \includegraphics[width=0.7\linewidth]{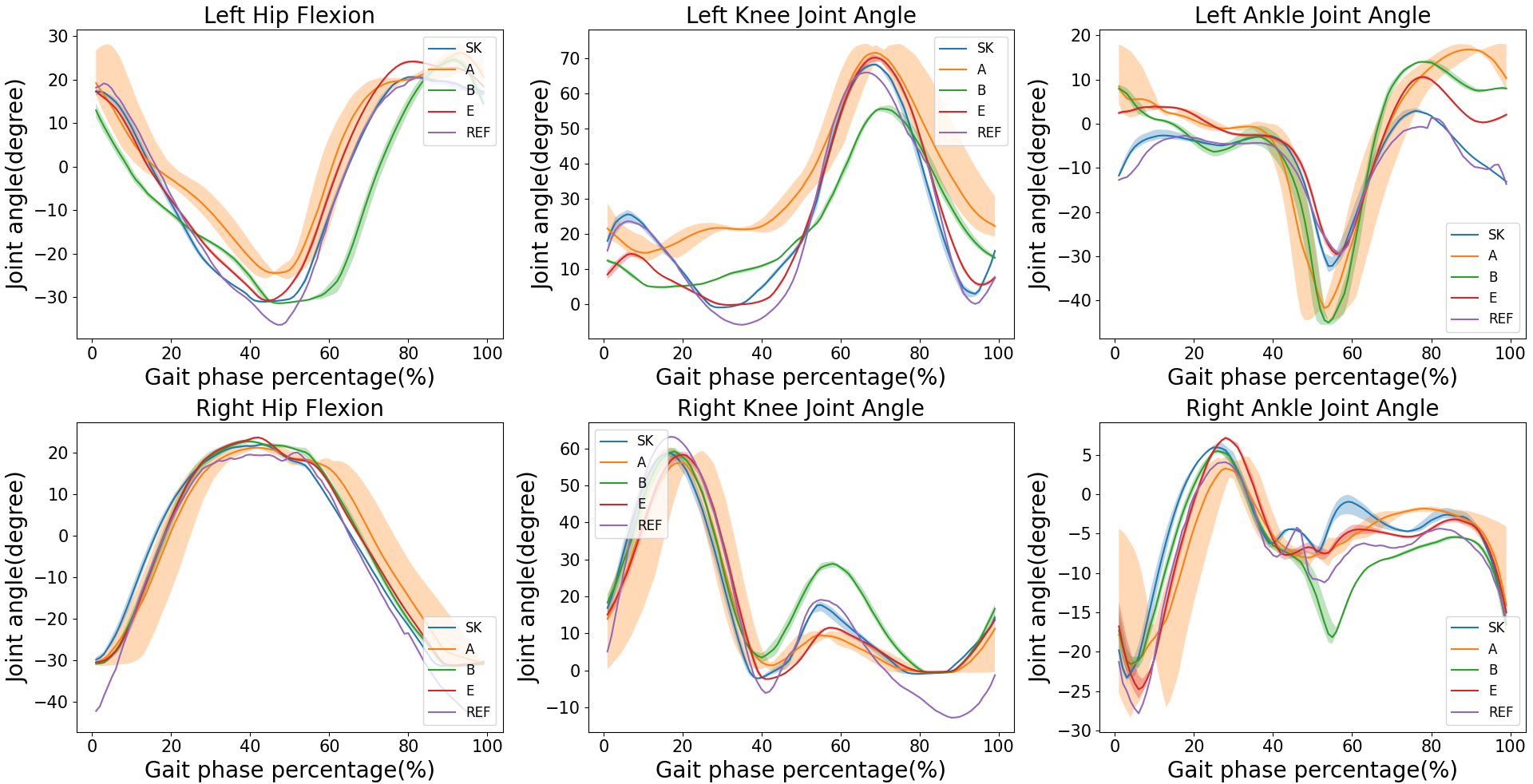}
    \caption{Joint angles for models SK, A, B, and E}
    \label{fig:joint_angles_1}
\end{figure}

The joint angles against the gait percentage were plotted for kinematic analysis of the walking control policies. Four complete gait cycles were extracted from each model from the simulation data. They were plotted as a solid line and shaded area, depicting the mean value and the distribution of values respectively. The reference motion, which the models were trained with, was plotted as a single curve. Figure \ref{fig:joint_angles_1} shows the joint angle plots for these models. 
All models appear successful in following the general profile of the reference motion (labelled as “REF”). For all joint angles, model SK appears to follow the reference motion better than models A and B. This might be due to the method of scaling and attaching the flex feet to the SK model, resulting in slight differences in the flex feet shape compared to the actual subject’s feet shape. Like the effects observed in the GRF plots, the slight difference in the flex feet shape may have affected the joint angle outputs, as the models may require slightly different joint angles to complete the same motion. In this case, the convex hulls of the SK feet might be more similar to the actual feet shape, particularly around the bottom of the feet, justifying why the model SK results are closest to the reference motion. 

Additionally, the joint angles of model A deviated quite significantly from this mean value, as seen from the orange shaded areas which show the distribution of its joint angle values. This result supports the hypothesis that the initial definition of the flex stiffness property is too low, resulting in inconsistency in the walking motion of model A. With the increasing stiffness of the flex, it can be seen that the distribution of the joint angles is much more consistent, especially with model B. With model E, where the position of the flex foot was adjusted slightly and the stiffness of flex material properties increased, we can see from the plots that there is an improvement in the joint angle results in model E over models A and B. This can be seen by the smaller deviation from the reference results, across all the different joint angles. 

Quantitative analysis of the joint angles is also performed using a Linear Fit Method (LFM), referenced from a study on determining waveform similarity from clinical gait data~\cite{ref_a}. A linear function, $Y_a$, is used to approximate the corresponding values from the experimental data based on the reference value $P_{ref}$.
\[Y_a=a_1*P_{ref}+a_0\]
 
Such an approach was chosen to determine how closely the experimental values matched the reference data. To determine how well the fitting line approximates the experimental values, the coefficient of determination $R^2$ value is determined. For an ideal result from experimental data, the result will be exactly equal to that of the reference data. For such a case we will get $R^2=1$, $a_1=1$, and $a_0=0$.

For the results of models SK, A, B, and E, the experimental values were fitted and $R^2$, $a_1$, and $a_0$ were determined, as shown in Table \ref{tab:joint_angle_1}.

\begin{table}
    \centering
\caption{$R^2$, $a_1$, and $a_0$ values for models SK, A, B, and E}
    \scriptsize
    \setlength{\tabcolsep}{2pt}
    \begin{tabular}{cccccccc} 
    \hline
    &  Mean ± SD&  L Hip&  L Knee&  L Ankle&  R Hip&  R Knee& R Ankle\\ 
         \hline
         &  $R^2$&  0.990±0.000&  0.978±0.007&  0.950±0.030&  0.988±0.008&  0.928±0.018& 0.893±0.063\\ 
         SK&  $a_1$&  0.955±0.005&  0.930±0.010&  1.090±0.050&  0.870±0.010&  0.778±0.018& 0.843±0.073\\ 
 & $a_0$& 0.000±0.010& 0.065±0.015& 0.030±0.000& 0.050±0.000& 0.093±0.008&0.033±0.008\\ 
  \hline
 & $R^2$
& 0.905±0.055& 0.730±0.190& 0.555±0.085& 0.923±0.093& 0.835±0.185&0.715±0.545\\ 
 A& $a_1$
& 0.808±0.053& 0.698±0.068& 1.658±0.103& 0.833±0.093& 0.743±0.113&0.798±0.408\\ 
 & $a_0$& 0.088±0.048& 0.338±0.053& 0.208±0.043& 0.063±0.053& 0.075±0.015&0.000±0.060\\ 
  \hline
 & $R^2$
& 0.818±0.038& 0.743±0.043& 0.695±0.045& 0.985±0.005& 0.953±0.008&0.805±0.015\\ 
 B& $a_1$
& 0.825±0.005& 0.625±0.015& 1.718±0.053& 0.905±0.015& 0.743±0.023&0.833±0.042\\ 
 & $a_0$& -0.083±0.007& 0.148±0.013& 0.188±0.023& 0.058±0.007& 0.178±0.007&-0.023±0.008\\
  \hline
 & $R^2$
& 0.978±0.008& 0.945±0.005& 0.838±0.008& 0.985±0.005& 0.918±0.007&0.945±0.035\\ 
 E& $a_1$
& 0.948±0.008& 0.988±0.018& 1.225±0.025& 0.895±0.005& 0.775±0.015&0.968±0.058\\ 
 & $a_0$& 0.040±0.000& 0.023±0.008& 0.138±0.008& 0.055±0.005& 0.085±0.005&0.023±0.008\\ 
    \end{tabular}
    \label{tab:joint_angle_1}
\end{table}
The results of the LFM are consistent with that from analysis of the plot. While all four models appear to show a linear relationship to the reference values, the joint angles from model SK are the closest to that of the reference when compared to models A and B, with consistent $R^2$ values approaching one, $a_1$ values approaching one, and $a_0$ values approaching zero for all six joint angles. We can also observe an improvement in the $R^2$, $a_1$, and $a_0$ values from models A to B, with a smaller deviation in these values as well, further strengthening the hypothesis that the initial definition of flex properties in model A had a stiffness value which was too low.

For the final model E, the adjustments of the flex foot position and increase of stiffness of the flex material properties have resulted in an improvement in the $R^2$, $a_1$, and $a_0$ values from models B to E, with these values approaching that of model SK. From this result, we may surmise that the approach to fine-tuning the stiffness and position of the flex feet was successful in better matching the joint angle results.

\section{Conclusion and Future Work}
In summary, the paper successfully used MuJoCo flex objects to generate shape-representative feet with soft-body capabilities, integrated them with a human SK model, and fine-tuned the position and flex material properties. The GRF results generated were comparable to the real GRF recorded, demonstrating the viability of soft-body simulations with flex in MuJoCo in approximating the soft tissues in humans. Hence, this presents an interesting avenue to produce a dynamically accurate human model in simulation driven by their own control policy while only seeing kinematic information during training. As this approach was implemented on an existing SK model in MuJoCo, this can likely be replicated with other existing MSK and SK models that use MuJoCo as their physics engine. In a clinical setting, this would prove to be useful in simulations of individual subjects, where an understanding of the GRF in different scenarios may be important.

However, this paper utilised simplified human foot shapes to facilitate easier computation. Moving forward, higher-resolution foot shapes can be used to better approximate the subject’s feet. Additionally, a full-body flex implementation could be developed, with flex objects for every body segment. This may be useful for simulations of the interactions between the human and assistive robot and has potential for contact based interaction simulation with assistive and rehabilitation robots. As these higher fidelity models will incur higher computational costs, there is also a need to understand the ideal fidelity of the human model that is still viable to compute for simulations.


\begin{credits}
\subsubsection{\ackname} This research work is supported by A*STAR under its National Robotics Programme (NRP) BAU grant, project - Assistive Robotics Programme (Award No: M22NBK0074). The authors thank Guan Ming Lim for his assistance in setting up the SMPL model and Lek Syn Lim for his assistance with data collection and expertise in GRF results. Participation in the motion capture was voluntary and informed consent was obtained. IRB-2018-04-014 governs the motion capture, and data access is governed by IRB-2022-713, approved by Nanyang Technological University's Institutional Review Board.

\end{credits}
%
%
%

\begin{thebibliography}{8}
\tiny
\bibitem{ref_1}
U. Martinez-Hernandez, B. Metcalfe, T. Assaf, L. Jabban, J. Male and D. Zhang, “Wearable assistive robotics: A perspective on current challenges and future trends,” Sensors \textbf{21}(20), 6751 (2021). \doi{10.3390/s21206751}
\bibitem{ref_2}
World Health Organization, “Ethical issues in patient safety research: Interpreting existing guidance,” \url{https://iris.who.int/handle/10665/85371}, last accessed 2023/9/13
\bibitem{ref_4}
AMMR, “Welcome to the AMMR documentation!,” \url{https://anyscript.org/ammr-doc/index.html\#}, last accessed 2023/9/14
\bibitem{ref_5}
A. Seth, J. L. Hicks, T. K. Uchida, A. Habib, C. L. Dembia, J. J. Dunne, C. F. Ong, M. S. DeMers, A. Rajagopal, M. Millard, S. R. Hammer, E. M. Arnold, J. R. Yong, S. K. Lakshmikanth, M. A. Sherman, J. P. Ku and S. L. Delp, “OpenSim: Simulating musculoskeletal dynamics and neuromuscular control to study human and animal movement,” PLOS Computational Biology \textbf{17}(7), (2018). \doi{10.1371/journal.pcbi.1006223} 
\bibitem{ref_6}
S. Christensen, X. Li and S. Bai, “Modeling and analysis of physical human-robot interaction of an upper body exoskeleton in assistive applications,” Modeling, Identification and Control: A Norwegian Research Bulletin \textbf{42}(4), 159–172 (2021). \doi{10.4173/mic.2021.4.2}
\bibitem{ref_7}
R. Ye, W. Xu, H. Fu, K. R. Jenamani, V. Nguyen, C. Lu, K. Dimitropoulou and T. Bhattacharjee, “RCare World: A human-centric simulation world for Caregiving Robots,” 2022 IEEE/RSJ International Conference on Intelligent Robots and Systems (IROS), 33-40 (2022). \doi{10.1109/IROS47612.2022.9982244} 
\bibitem{ref_11}
Z. Q. Cheng, Y. Chen, R. R. Martin, T. Wu and Z. Song, “Parametric modeling of 3D human body shape—A survey,” Computers \& Graphics \textbf{71}, 88-100 (2018). \doi{10.1016/j.cag.2017.11.008}
\bibitem{ref_xpbd}
M. Macklin,  M. Müller and N. Chentanez, “XPBD: Position-Based Simulation of Compliant Constrained Dynamics,” Motion In Games, 49-54 (2016). \doi{10.1145/2994258.2994272}
\bibitem{ref_26}
K. K. Dwivedi, P. Lakhani, S. Kumar and N. Kumar, “A hyperelastic model to capture the mechanical behaviour and histological aspects of the soft tissues,” Journal of the Mechanical Behavior of Biomedical Materials \textbf{126}, 105013 (2022). \doi{10.1016/j.jmbbm.2021.105013}
\bibitem{ref_beth}
W. T. Ang, "Investigation of Modeling Differences between OpenSim and Visual3D for Gait Analysis of Healthy Gait," 16th International Convention on Rehabilitation Engineering and Assistive Technology (i-CREATe '23), Pathum Thani, Thailand, pp. 42-46 (2023). \doi{10.1145/3628228.3628492} 
\bibitem{ref_mocap}
P. Liang, W. H. Kwong, A. Sidarta, C. K. Yap, W. K. Tan, L. S. Lim, P. Y. Chan, C. K. K. Wee, S. K. Wee, K. Chua, C. Quek and W. T. Ang, "An Asian-centric human movement database capturing activities of daily living," Scientific Data \textbf{7}, 290 (2020). \doi{10.1038/s41597-020-00627-7}
\bibitem{ref_rris}
S. S. Chan, M. Lei, H. Johan, and W. T. Ang, "Creation and Evaluation of Human Models with Varied Walking Ability from Motion Capture for Assistive Device Development," 2023 International Conference on Rehabilitation Robotics (ICORR), Singapore, pp. 1-6. IEEE (2023). \doi{10.1109/ICORR58425.2023.10304741}
\bibitem{ref_smpl}
M. Loper, N. Mahmood, J. Romero, G. Pons-Moll and M. J. Black, “SMPL: A Skinned Multi-Person Linear Model,” ACM Transition on Graphic \textbf{34}(6), pp. 1-16 (2015). \doi{10.1145/2816795.2818013} 
\bibitem{ref_22}
P. Arpaia, D. Cuneo, S. Grassini and F. Mancino, “A finite element model of abdominal human tissue for improving the accuracy in insulin absorption assessment: A feasibility study,” Measurement Sensors \textbf{18}(11), 100218 (2021). \doi{10.1016/j.measen.2021.100218}
\bibitem{ref_9}
DeepMind, “MuJoCo Documentation: Overview,” \url{https://mujoco.readthedocs.io/en/stable/overview.html}, last accessed 2023/9/23
\bibitem{ref_27}
G. Singh and A. Chanda, “Mechanical properties of whole-body soft human tissues: a review,” Biomedical Materials \textbf{16}(6), (2021). \doi{10.1088/1748-605X/ac2b7a}
\bibitem{ref_28}
N. Arnold, J. Scott and T. R. Bush, “A review of the characterizations of soft tissues used in human body modeling: Scope, limitations, and the path forward,” Journal of Tissue Viability, \textbf{32}(2), 286-304 (2023). \doi{10.1016/j.jtv.2023.02.003}
\bibitem{ref_29}
R. Bonnaire, W.-S. Han, P. Calmels, R. Convert and J. Molimard, "Parametric study of lumbar belts in the case of low back pain: effect of patients' specific characteristics," Computational Biomechanics for Medicine, 43-59 (2020). \doi{10.1007/978-3-030-15923-8_4}
\bibitem{ref_30}
T. Jadidi, M. R. Tabar and A. Mashaghi, “Poisson’s ratio and Young’s modulus of lipid bilayers in different phases,” Frontiers in Bioengineering and Biotechnology \textbf{2}(22), (2014). \doi{10.3389/fbioe.2014.00008} 
\bibitem{ref_34}
P. Schumacher, T. Geijtenbeek, V. Caggiano, V. Kumar, S. Schmitt, G. Martius and D. F. B. Haeufle, “Natural and Robust Walking using Reinforcement Learning without Demonstrations in High-Dimensional Musculoskeletal Models,” arXiv, (2023). \doi{10.48550/arXiv.2309.02976}
\bibitem{ref_a}
M. Iosa, A. Cereatti, A. Merlo, I. Campanini, S. Paolucci, and A. Cappozzo, "Assessment of Waveform Similarity in Clinical Gait Data: The Linear Fit Method," Biomed Res Int, 214156 (2014). \doi{10.1155/2014/214156}



\end{thebibliography}
%

\end{document}